\documentclass{ecai} 

\usepackage{latexsym}
\usepackage{amssymb}
\usepackage{amsmath}
\usepackage{amsthm}
\usepackage{booktabs}
\usepackage{enumitem}
\usepackage{graphicx}
\usepackage{color}
\usepackage{listings}
\usepackage{algorithm}
\usepackage{algpseudocode}

\newcommand{\BibTeX}{B\kern-.05em{\sc i\kern-.025em b}\kern-.08em\TeX}

\lstdefinestyle{mystyle}{        
    keepspaces=false,                                  
    showspaces=false,                
    showstringspaces=false,
    showtabs=false,                  
    tabsize=4,
    basicstyle=\footnotesize
}

\begin{document}


\begin{frontmatter}


\paperid{123} 


\title{Program Synthesis using Inductive Logic Programming for the Abstraction and Reasoning Corpus}


\author[A]{\fnms{Filipe}~\snm{Marinho Rocha}}
\author[A]{\fnms{Inês}~\snm{Dutra}}
\author[A]{\fnms{Vítor}~\snm{Santos Costa}}

\address[A]{INESCTEC-FCUP}


\begin{abstract}
The Abstraction and Reasoning Corpus (ARC) is a general artificial intelligence benchmark that is currently unsolvable by any Machine Learning method, including Large Language Models (LLMs).
It demands strong generalization and reasoning capabilities which are known to be weaknesses of Neural Network based systems. In this work, we propose a Program Synthesis system that uses Inductive Logic Programming (ILP), a branch of Symbolic AI, to solve ARC. We have manually defined a simple Domain Specific Language (DSL) that corresponds to a small set of object-centric abstractions relevant to ARC. This is the Background Knowledge used by ILP to create Logic Programs that provide reasoning capabilities to our system. The full system is capable of generalize to unseen tasks, since ILP can create Logic Program(s) from few examples, in the case of ARC: pairs of Input-Output grids examples for each task.
These Logic Programs are able to generate Objects present in the Output grid and the combination of these can form a complete program that transforms an Input grid into an Output grid.
We randomly chose some tasks from ARC that don't require more than the small number of the Object primitives we implemented and show that given only these, our system can solve tasks that require each, such different reasoning.

\end{abstract}

\end{frontmatter}


\section{Introduction}

Machine Learning\cite{carbonell1983overview}, more specifically, Deep Learning\cite{lecun2015deep}, has achieved great successes and surpassed human performance in several fields. These successes, though, are in what is called skill-based or narrow AI, since each DL model is prepared to solve a specific task very well but fails at solving different kind of tasks\cite{marcus2018deep}\cite{lake2017building}.

It is known Artificial Neural Networks (ANNs) and Deep Learning (DL) suffer from lack of generalization capabilities. Their performance degrade when they are applied to Out-of-Distribution data \cite{kirchheim2024out} \cite{farquhar2022out} \cite{ye2022ood}.

Large Language Models (LLMs), more recently, have shown amazing capabilities, shortening the gap between Machine and Human Intelligence. But they still show lack of reasoning capabilities and require lots of data and computation.

The ARC challenge was designed by Francois Chollet in order to evaluate whether our artificial intelligent systems have progressed at emulating human like form of general intelligence. ARC can be seen as a general artificial intelligence benchmark, a program synthesis benchmark, or a psychometric intelligence test \cite{chollet2019measure}. 

It was presented in 2019 but it still remains an unsolved challenge, and even the best DL models, such as LLMs cannot solve it \cite{lee2024reasoning}\cite{butt2024codeit}\cite{bober2024neural}. GPT-4V, which is GPT4 enhanced for visual tasks \cite{achiam2023gpt}, is unable to solve it too \cite{xu2023llms} \cite{mitchell2023comparing} \cite{singh2023assessing}.

It targets both humans and artificially intelligent systems and aims to emulate a human-like form of general fluid intelligence. It is somewhat similar in format to Raven’s Progressive Matrices \cite{raven2003raven}, a classic IQ test format.

It requires, what Chollet describes as, developer-aware generalization, which is a stronger form of generalization, than Out-of-Distribution generalization \cite{chollet2019measure}. In ARC, the Evaluation set, with 400 examples, only features tasks that do not appear in the Training set, with also 400 examples, and all each of these tasks require very different Logical patterns to solve, that the developer cannot foresee. There is also a Test set with 200 examples, which is completely private.

For a researcher setting out to solve ARC, it is perhaps best understood as a program
synthesis benchmark \cite{chollet2019measure}. Program synthesis \cite{gulwani2015inductive} \cite{gulwani2017program} is a subfield of AI with the purpose of generating of programs that satisfy a high-level specification, often provided in the form of example pairs of inputs and outputs for the program, which is exactly the ARC format.

Chollet recommends starting by developing a domain-specific language (DSL) capable of expressing all possible solution programs for any ARC task\cite{chollet2019measure}. Since the exact set of ARC tasks is purposely not formally definable, and can be anything that would only involve Core Knowledge priors, this is challenging.

Objectness is considered one of the Core Knowledge prior of humans, necessary to solve ARC \cite{chollet2019measure}. Object-centric abstractions enable object awareness which seems crucial for humans when solving ARC tasks \cite{acquaviva2022communicating}\cite{johnson2021fast} and is central to general human visual understanding \cite{spelke2007core}. There is previous work using object-centric approaches to ARC that shows its usefulness \cite{lei2024generalized} \cite{acquaviva2022communicating}.

Inductive Logic Programming (ILP)\cite{muggleton1994inductive} is also considered a Machine Learning method, but to our knowledge it was never applied to the ARC challenge. It can perform Program Synthesis \cite{cropper2022inductive} and is known for being able to learn and generalize from few training examples \cite{lin2014bias}\cite{muggleton2018meta}.

We developed a Program Synthesis system that uses ILP, on top of Object-centric abstractions, manually defined by us. It does Program synthesis by searching the combination of Logic relations between objects existent in the training examples. This Logic relations are defined by Logic Programs, obtained using ILP.

The full program our system builds, is composed by Logic Programs that are capable of generating objects in the Output Grid.

We selected five random examples that contain only simple geometrical objects and applied our system to these.

\begin{figure}[h]
\centering
\includegraphics[width=9 cm]{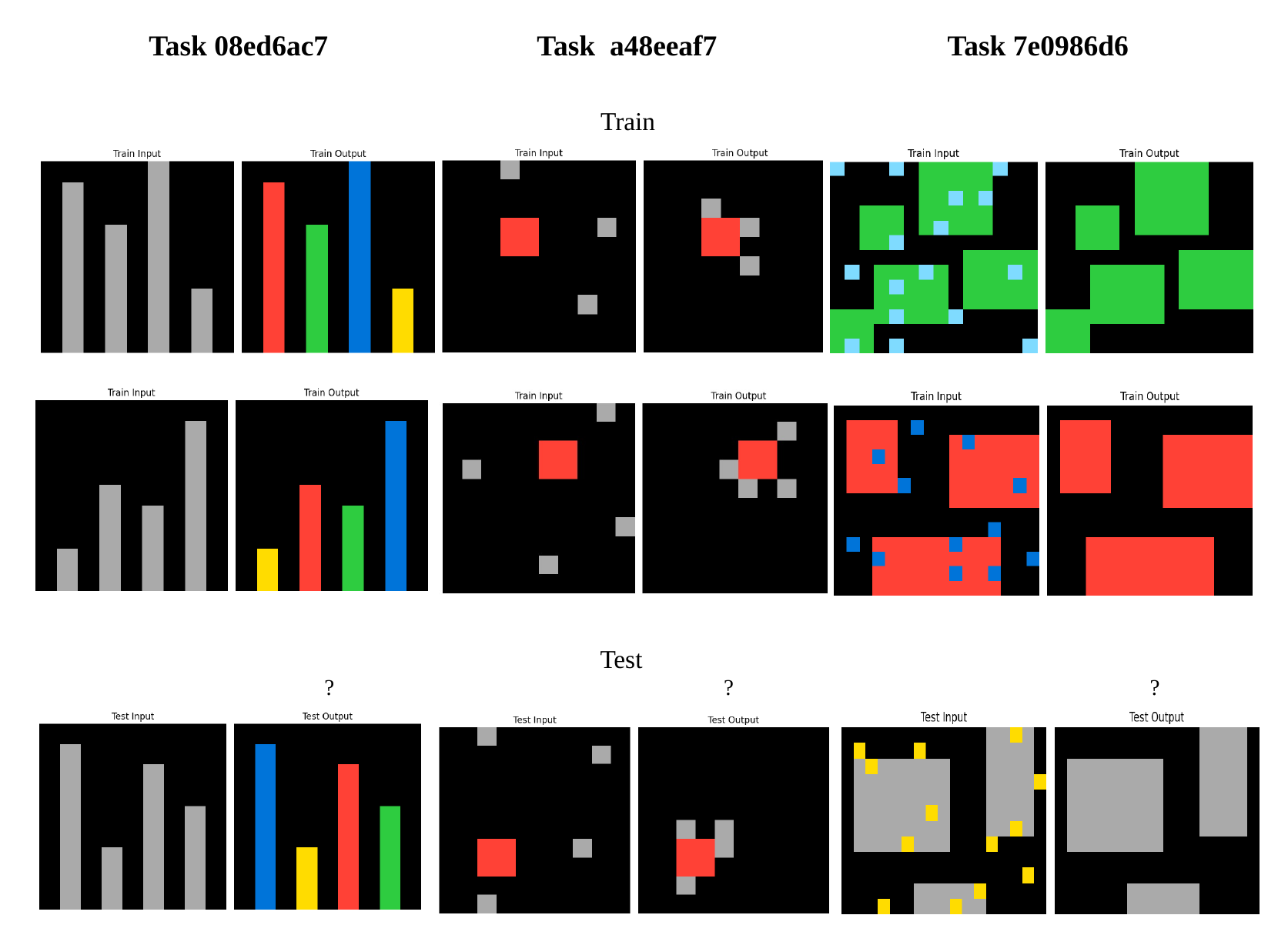}
\caption{Example tasks of the ARC Training dataset with the solutions shown. The goal is to produce the Test Output grid given the Test Input grid and the Train Input-Output grid pairs. 
We can see the logic behind each task is very different and the relations between objects are key to the solutions.}
\label{fig:example_tasks}
\end{figure}


\section{Object-centric Abstractions and Representations}

Object-centric abstractions reduce substantially the search space by enabling the focus on the relations between objects, instead of individual pixels.

However, there may be multiple ways to interpret the same image in terms of objects, therefore, we keep multiple, overlapping Object representations for the same image.


\subsection{Objects and Relations between Objects}

We have defined manually a simple DSL that is composed by the Objects: Point, Line and Rectangle and the Relations between Objects:  LineFromPoint, Translate, Copy, PointStraightPathTo.

\begin{table}[h]
\caption{Object types.}
\centering
\begin{tabular}{ll@{\hspace{8mm}}ll}
\hline
Objects & Attributes  \\
\toprule
Point & (x, y, color)  \\
Line & (x1, y1, x2, y2, color, len, orientation, direction)  \\
Rectangle & (x1, y1, x2, y2, x3, y3, x4, y4, color, clean, area) \\
\bottomrule
\end{tabular}
\end{table}

\begin{table}[h]
\caption{Relations types.}
\centering
\begin{tabular}{ll@{\hspace{8mm}}ll} 
 \hline
Relations & Attributes  \\

\toprule
LineFromPoint & (point, line, len, orientation, direction) \\
Translate & (obj1, obj2, xdir, ydir, color2) \\
Copy & (obj1, obj2, color2, clean) \\
PointStraightPathTo & (point, obj, xdir, ydir, orientation, direction) \\
\bottomrule
\end{tabular}
\end{table}


\subsection{Multiple Representations}

An image representation in our object-centric approach is defined by a list of objects (possibly overlapping) and a background color. This representation can build a target image from an empty grid by firstly filling the grid with the background color and then drawing each object on top of it.

An image grid can be defined by multiple Object representations. For example a single Rectangle in an empty grid can also be defined as several Line or Point objects that form the same Rectangle. 

Since we don't know from the start, the Logic behind the transformation of the Input grid into the Output grid, we don't know if the Object Rectangle is key to this logic (relation between rectangles) or if instead the logic demands using a Line representation for it may involve a relation dependent on lines, such as the LineFromPoint relation. 

\begin{figure}[h]
\centering
\includegraphics[width=6 cm]{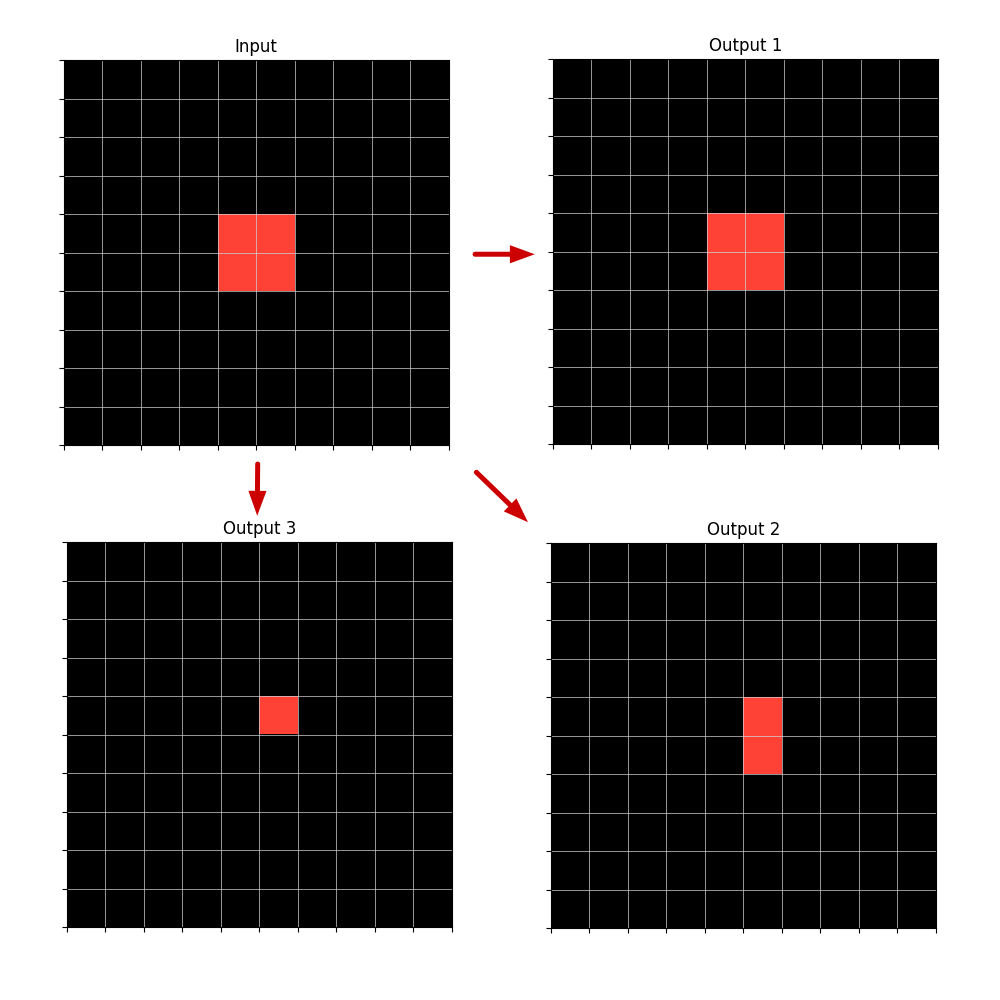}
\caption{Example of an Input grid with a Rectangle (or with several contiguous Points or Lines) and three possible Output grids, built from the Input, depending on which type of Object representation is used. The Output object can be described by the relation Copy applied to the Input Rectangle or to just one Input Point or Input Line that is part of the same Rectangle.}
\label{fig:example_tasks}
\end{figure}

Likewise, the same image transformation can be explained by different relations.

So we work with multiple and intermingled representations of objects and relations until we get the final program or programs that can transform each of the Training input grids into the output grids and also produce a valid Output grid for the Test Example, which will be the output solution given by our system.

If multiple programs applied separately can produce successfully the same Input-Output Train images transformation, we can use any of this, or select one, for example: the shortest program, which will have more probability of being the correct one, according to the Occam principle.


\section{ILP}

Inductive Logic Programming (ILP) is a form of logic-based Machine Learning. The goal is
to induce a hypothesis, a Logic Program or set of Logical Rules, that generalizes given training examples and Background Knowledge (BK). Our DSL composed by Objects and Relations, is the BK given.

As with other forms of ML, the goal is to induce a
hypothesis that generalizes training examples. However, whereas most forms of ML use
vectors/tensors to represent data, ILP uses logic programs. And
whereas most forms of ML learn functions, ILP learns relations.

The fundamental ILP problem is to efficiently search a large hypothesis space. There are
ILP approaches that search in either a top-down or bottom-up fashion and others that combine both. We use a top-down approach in our system.

Learning a large program with ILP is very challenging, since the search space can get very big. There are some approaches that try to overcome this.

\subsection{Divide-and-conquer}

Divide-and-conquer approaches \cite{witt2023divide} divide
the examples into subsets and search for a program for each subset.
We use a Divide-and-conquer approach but instead of applying it to examples, we apply it to the Objects inside the examples.


\section{Program Synthesis using ILP}
ILP is usually seen as a method for Concept Learning but is capable of building a Logic Program in Prolog, which is Turing-complete, hence, it does Program Synthesis.
We extend this by combining Logic Programs to build a bigger program that generates objects in sequence, to fill an empty grid, which corresponds to, procedurally, applying grids transformations to reach the solution.

A Relation can be used to generate objects. All of our Relations, except for PointStraightPathTo, are able to generate Objects given the  first Object of the Relation. To generate an unambiguous output, the Relation should be defined by a Logic Program. This Logic Program is built by using ILP in our system.

Example of a logic program in Prolog:

\begin{lstlisting}[language=Prolog]

line_from_point(Point,Line,Len,Orientation,Direction):- 
	member(Point,Input_points), 
	equal(Len,5), 
	equal(Orientation,'vertical').
	
\end{lstlisting}

\begin{figure}[h]
\centering
\includegraphics[width=6 cm]{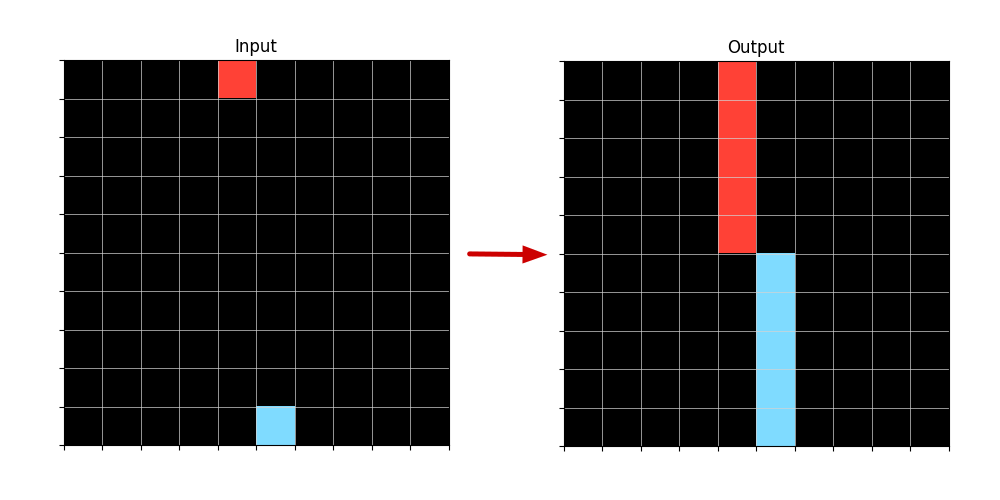}
\caption{Example of an Input-Output transformation when applied a LineFromPoint Logic Program that is able to generate Lines from Points.}
\label{fig:line_from_point_example1}
\end{figure}

This Logic Program can generate unambiguously, two lines in the Output. The variable Direction is not needed for this, since the points are on the edges of the Grid and so, can only grow into lines in one Direction each.

If the program was shorter, it would generate multiple Lines from each Input Point. Example of the same Logic Program without the last term in the body:

\begin{lstlisting}[language=Prolog]

line_from_point(Point,Line,Len,Orientation,Direction):- 
	member(Point,Input_points), 
	equal(Len,5).
	
\end{lstlisting}

\begin{figure}[h]
\centering
\includegraphics[width=6 cm]{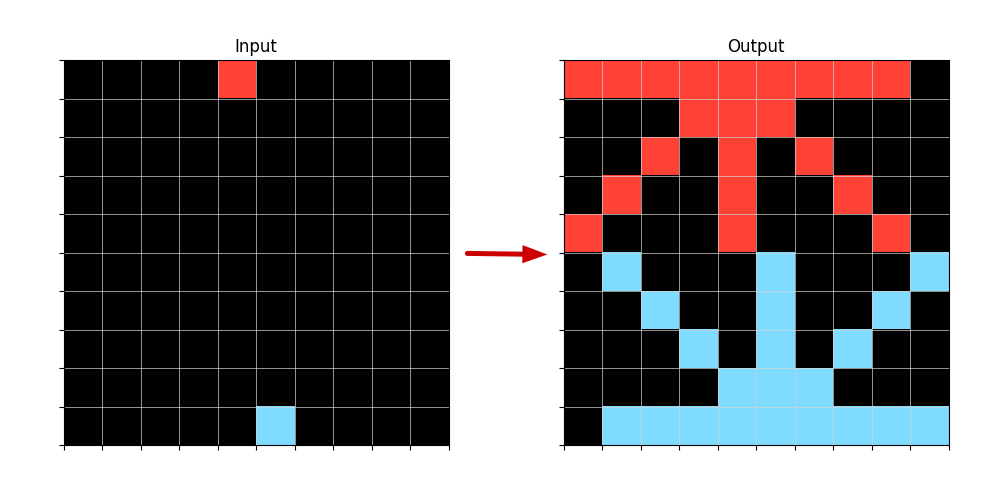}
\caption{Example of an Input-Output transformation when applied a LineFromPoint Logic Program without having the Orientation defined.}
\label{fig:line_from_point_example1}
\end{figure}

So by obtaining Logic Programs through ILP we are indeed constructing a program that generates objects that can fill an empty Test Output grid, in order to reach the solution.


\section{System Overview}
\subsection{Objects and Relations Retrieval}
Our system begins by retrieving all Objects defined in our DSL, in the Input and Output grids for each example of a task.
As we reported before, we keep multiple object representations that may overlap in terms of occupying the same pixels in a image. 

Then we search for Relations defined in our DSL betweeen the found objects. We search for Relations between Objects only present in the Input Grid: Input-Input Relations, only present in the Output Grid: Output-Output Relations and between Objects in the Input Grid and Output Grid: Input-Output Relations.

The type of Objects previously found can constraint the search for Relations, since some Relations are specific to some kind of Objects.

\subsection{ILP calls}
We then call ILP to create Logic Programs to define only the Relations found in the previous step. This also reduces the space of the search.
We start with the Input-Output Relations, since we need to build the Output objects using information from the Input. After we generate some Object(s) in the Output grid we can starting using also Output-Output relations to generate Output Objects from other Output Objects.
Input-Input relations can appear in the body of the rules but won't be the Target Relations to be defined by ILP, since they don't generate any Object in the Output.

After each ILP call what is considered Input information, increases. The relations and Objects returned by the ILP call that produce an updated Grid, are considered now as it were present in the Input, and this updated Grid is taken as the initial Output grid where to build on, in the subsequent ILP calls.

\subsubsection{Candidate Generation}
The candidates terms to be added to the Body of the Rule (Horn Clause) are the Objects and Relations found in the Retrieval step (5.1) plus the Equal(X,...), GreaterThan(X,...), LowerThan(X,...), Member(X,...) predicates. GreaterThan and Lowerthan relate only number variables. We also consider aX+b , being X a number variable and a and b, constants in some interval we predefine.

Since we are using Typed Objects and Relations we only need to generate those candidates that are related to each Target Relation variable by type.

For example, in building a Logic Program that defines the Relation:
LineFromPoint(point,line,len,orientation,direction)
we are going to generate candidates that relate Points or attributes of Points in order to instantiate the first variable Point of the relation.
Then we proceed to the other variables (besides the line variable): len, orientation and direction. Len is of type Int, so it is only relevant to be related to other Int variables or constants.
The variable Line remains free because it is the Object we want the relation to generate.

\subsubsection{Positive and Negative examples}

As we have seen in Section 3, FOIL requires positive and negative examples to induce a program.
ARC dataset is only composed of Positive examples. So we created a way to represent negative examples.

In our system for each Logic Program the Positive examples are the Objects that this program generates that exist in the Training Data, and the Negative examples are the Objects that the program generates, but don't exist in the Training Data. 

For example, imagine the Ouput in Figure 3 is the correct one, but the program generates all the Lines in Figure 4 like the shorter Prolog program we presented. For this Program the positives it covers would be the two Lines in Figure 3 and the negatives covered would be all the Lines in Figure 4, except for the vertical ones. 

Here, our Object-centric approach also reduces the complexity of the search, compared with taking the entire Grid into account, to define what is a positive example or not.

\subsubsection{Unification between Training examples}
An ILP call is made for each Relation and all of the Task's Training examples. Our ILP system is constrained to produce Programs that can unify between examples, that is, they are abstract and generalize to two or more of the training examples.

Why two and not all of the training examples? One of the five examples we selected to help us develop our system showed us that a program that solves all Training examples can be too complex and not necessary to solve the Test example.

\begin{figure}[h]
\centering
\includegraphics[width=6 cm]{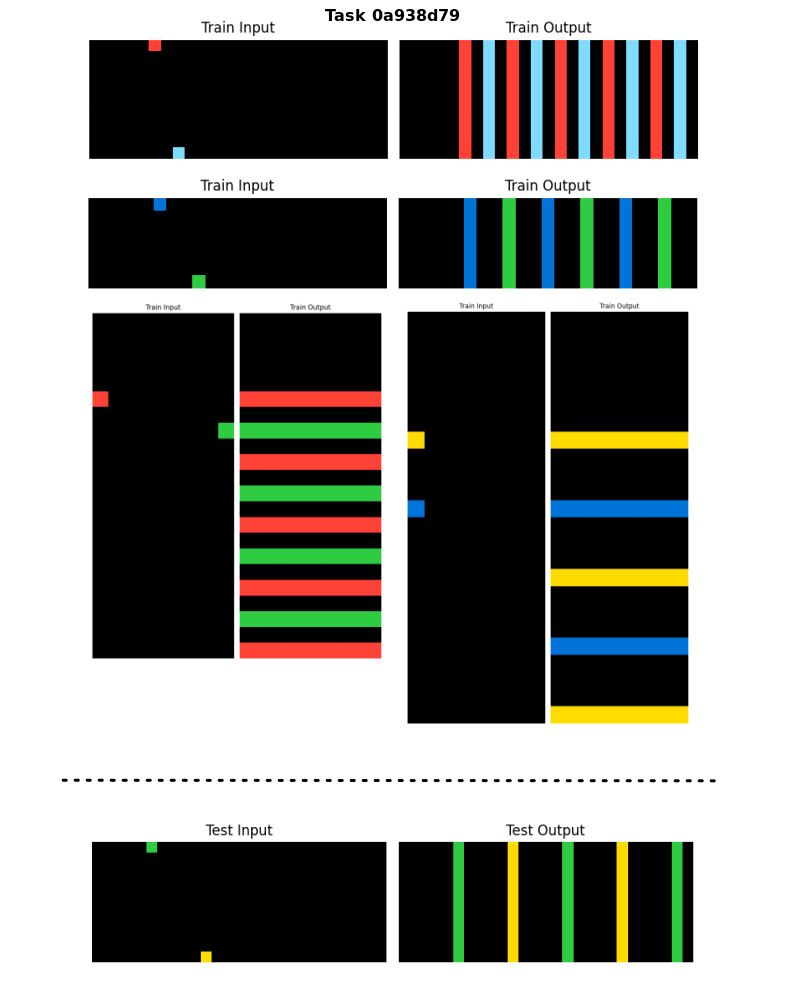}
\caption{Example task that contains two Train examples with vertical lines and two Train examples with horizontal lines, in the Output grids. The Test example only requires vertical Lines in the same way as the first two Train examples.}
\label{fig:example_unification}
\end{figure}

In Figure 5 we can see a sample task and derive the logic for its solution: draw Lines from Points until the opposite border of the grid and then Translate these Lines repeatedly in the perpendicular direction of the Lines until the end of the grid.

To unify the four Train examples in this task and following the logic we described for this solution, it would require a more complex program than a solution that only unifies the first two examples.

Let's see one sequence of Logic programs that solves the first two Train examples, and it would also solve, sucessfully, the Test example:

\begin{lstlisting}[language=Prolog]

line_from_point(Point,Line,Len,Orientation,Direction):- 
	member(Point,Input_points), 
	equal(Len,X_dim), 
	equal(Orientation,'vertical').
	
translate(Line1,Line2,X_dir,Y_dir):-
	member(Line1,Input_lines),
	equal(X_dir,0),
	translate(Input_point1,Input_point2,X_dir,Y_dir),
	equal(Y_dir,2*Y_dir).
	
				...
				
  (the same translate program three more times)			

\end{lstlisting}

The sequence of Logic Programs that would solve the last two Train examples with the Horizontal lines, but wouldn't solve sucessfully the Test example which has Vertical lines:

\begin{lstlisting}[language=Prolog]

line_from_point(Point,Line,Len,Orientation,Direction):- 
	member(Point,Input_points), 
	equal(Len,Y_dim), 
	equal(Orientation,'horizontal').
	
translate(Line1,Line2,X_dir,Y_dir):-
	member(Line1,Input_lines),
	equal(Y_dir,0),
	translate(Input_point1,Input_point2,X_dir,Y_dir),
	equal(X_dir,2*X_dir).
	
				...
				
  (the same translate program three more times)			

\end{lstlisting}

The Test example only needs two translations in the Output Grid, so a program with more translations would work, since it would fill the entire grid and the extra translations just wouldn't apply. Our system consider this a valid program.

But if the Test grid was longer and required more translations than present in the Training examples, our program wouldn't work, since the number of translations wouldn't produce the exact solution, but an incomplete one. For this kind of tasks we would need to use higher-order constructs as: Do Until, Repeat While or Recursion with a condition, to apply the same Relation a number of times or until some condition fails or is triggered. This is scope of future work.

\subsection{Rules, Grid states and Search}

The first ILP call is made on empty grid states.
Each ILP call produces a rule that defines a Relation that generates objects, and returns new Output grid states with these objects added, for each example and the Test example. Each subsequent ILP call is made on the updated Output Grid states.

The search will be on finding the right sequence of Logic Programs that can build the Output Grids, starting from the empty grid. When we have one Program that builds at least one complete Training example Output Grid, consisting of a program that is unified between two or more examples and can also produce a valid solution (not necessarily the correct one) in the Test Output Grid, we consider it the final program.

A valid program is one that builds a consistent theory. For example, a program that generates different colored Lines that intersect with each other, is inconsistent, since both Lines cannot exist in the Output grid (one overlaps the other). We consider programs like this invalid, and discard them, when searching for the complete program.

\begin{figure}[h]
\centering
\includegraphics[width=5 cm]{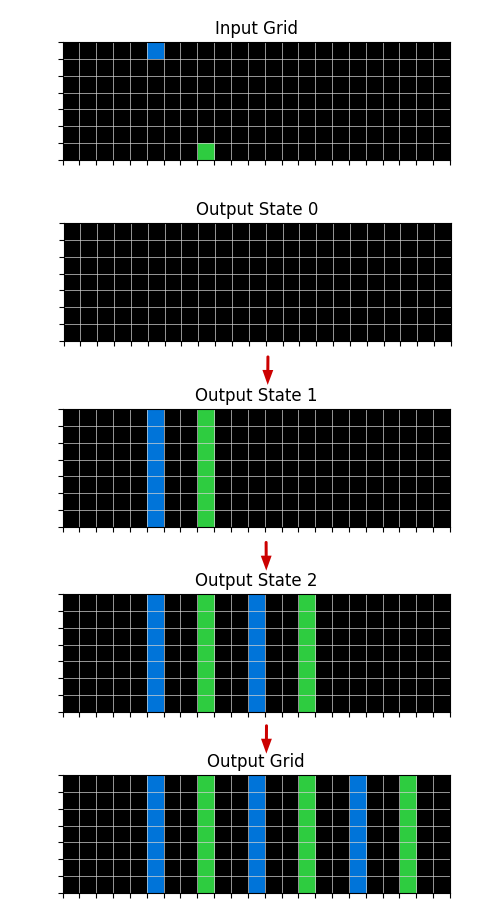}
\caption{Example of a sucessful sequence of transformations to the Output grid state reaching the final state which is the correct Output grid. These transformations correspond to a sequence of Logic Programs: first transition comes from LineFromPoint Input-Output program, the second and third transitions correspond to the Translate Output-Output programs.}
\label{fig:output_state_transition}
\end{figure}

\subsection{Deductive Search}

In Figure 7 we can see an example Task that in order to be solved, it would be easier to do it in reverse, that is, the Input Grid being generated from Output information, but since we cannot solve the Test Output Grid by working in this direction, we use a form of Deductive Search to overcome this.

When we apply a Logic Program constructed by ILP, we can have different results depending on the order of the Object Generation. When an Object is generated in the Output Grid, none other Object generated afterwards, can intersect with it. 

So the coverage of grid space of the same Logic Program may vary depending on the order of the program application. It is the procedural aspect of our system. When applying several Logic Programs in sequence, this problem gets even bigger.

So when applying the full program to produce the Test Output grid, we use Deductive search to apply the whole program in the way that covers the most surface. Since the final program is one that can cover the whole surface of Train Output grids, we should have a solution that can cover all of the Test Output grid too.

\begin{figure}[h]
\centering
\includegraphics[width=9 cm]{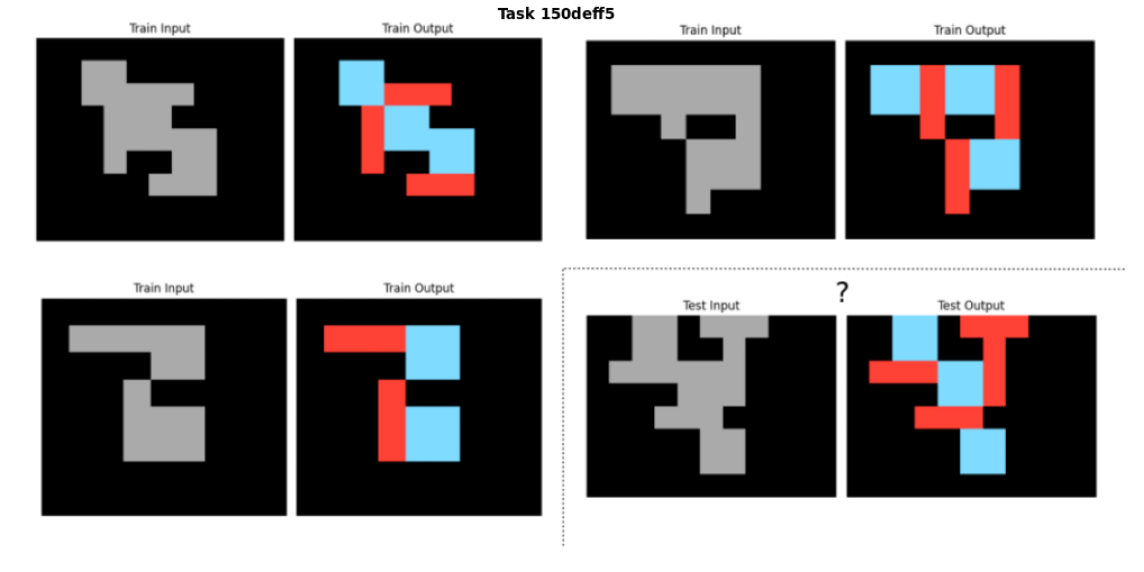}
\caption{Example task where the Output Grids are more informative than the Input Grids, to define and apply the Object Relations for the solution.}
\label{fig:last_sample_task}
\end{figure}


\section{Experiments}

Our system was applied sucessfully to five tasks, the three tasks in Figure 1: 08ed6ac7, a48eeaf7, 7e0986d6, the task in Figure 5: 0a938d79 and the task in Figure 7: 150deff5.

In the Appendices we present the Output solutions in Prolog for each task.


\section{Conclusion}

We showed our system is able to solve the 5 sample tasks selected. When we finish our software implementation we will apply our system to the full Training and Evaluation datasets.

ILP is at the core of our system and we showed that only by providing it with a small DSL or Background Knowledge, ILP is able to construct and represent the Logic behind solutions of ARC tasks.

ILP gives our system abstract learning and generalization capabilities, which is at the core of the ARC challenge.

Since the other ARC tasks may depend on many different DSL primitives, we plan to develop a way to automate the DSL creation.

As we mentioned before, potentially, we are going to need higher-order constructs to solve other tasks and plan to incorporate this into our system.




\bibliographystyle{unsrt}
\bibliography{paperV1}


\section*{Appendices}

\begin{figure}[h]
\centering
\includegraphics[width=8 cm]{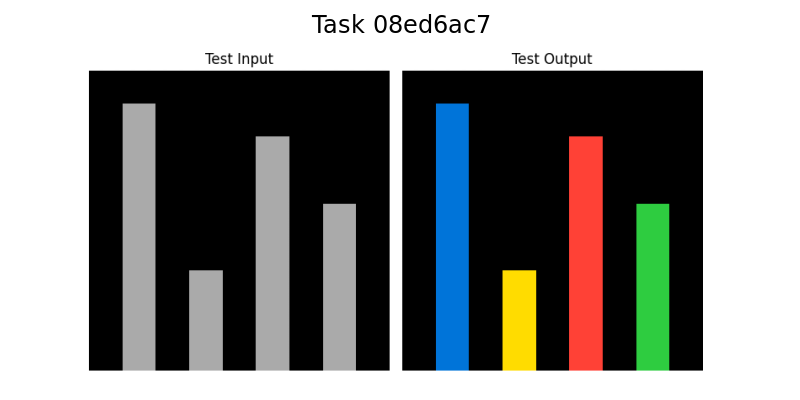}
\label{fig:08ed6ac7}
\end{figure}

\begin{lstlisting}[language=Prolog]
copy(Line1,Line_out,Color_out,Clean):-
	member(Line1,Input_lines),
	member(Line2,Input_lines),
	member(Line3,Input_lines),
	member(Line4,Input_lines),
	line_attrs(Line1,X11,Y11,X12,Y12,Color1,Len1,
		Orientation1,Direction1),
	line_attrs(Line2,X21,Y21,X22,Y22,Color2,Len2,
		Orientation2,Direction2),
	line_attrs(Line3,X31,Y31,X32,Y32,Color3,Len3,
		Orientation3,Direction3),
	line_attrs(Line4,X41,Y41,X42,Y42,Color4,Len4,
		Orientation4,Direction4),
	equal(Color_out,'blue'),
	equal(Clean,100),
	greaterthan(Len1,Len2), 
	greaterthan(Len1,Len3), 
	greaterthan(Len1,Len4).
	
copy(Line1,Line_out,Color_out,Clean):-
	member(Line1,Input_lines),
	member(Line2,Input_lines),
	member(Line3,Input_lines),
	member(Line4,Input_lines),
	line_attrs(Line1,X11,Y11,X12,Y12,Color1,Len1,
		Orientation1,Direction1),
	line_attrs(Line2,X21,Y21,X22,Y22,Color2,Len2,
		Orientation2,Direction2),
	line_attrs(Line3,X31,Y31,X32,Y32,Color3,Len3,
		Orientation3,Direction3),
	line_attrs(Line4,X41,Y41,X42,Y42,Color4,Len4,
		Orientation4,Direction4),
	equal(Color_out,'red'),
	equal(Clean,100),
	lowerthan(Len1,Len2), 
	greaterthan(Len1,Len3), 
	greaterthan(Len1,Len4).
	
copy(Line1,Line_out,Color_out,Clean):-
	member(Line1,Input_lines),
	member(Line2,Input_lines),
	member(Line3,Input_lines),
	member(Line4,Input_lines),
	line_attrs(Line1,X11,Y11,X12,Y12,Color1,Len1,
		Orientation1,Direction1),
	line_attrs(Line2,X21,Y21,X22,Y22,Color2,Len2,
		Orientation2,Direction2),
	line_attrs(Line3,X31,Y31,X32,Y32,Color3,Len3,
		Orientation3,Direction3),
	line_attrs(Line4,X41,Y41,X42,Y42,Color4,Len4,
		Orientation4,Direction4),
	equal(Color_out,'green'),
	equal(Clean,100),
	lowerthan(Len1,Len2), 
	lowerthan(Len1,Len3), 
	greaterthan(Len1,Len4).
	
copy(Line1,Line_out,Color_out,Clean):-
	member(Line1,Input_lines),
	member(Line2,Input_lines),
	member(Line3,Input_lines),
	member(Line4,Input_lines),
	line_attrs(Line1,X11,Y11,X12,Y12,Color1,Len1,
		Orientation1,Direction1),
	line_attrs(Line2,X21,Y21,X22,Y22,Color2,Len2,
		Orientation2,Direction2),
	line_attrs(Line3,X31,Y31,X32,Y32,Color3,Len3,
		Orientation3,Direction3),
	line_attrs(Line4,X41,Y41,X42,Y42,Color4,Len4,
		Orientation4,Direction4),
	equal(Color_out,'yellow'),
	equal(Clean,100),
	lowerthan(Len1,Len2), 
	lowerthan(Len1,Len3), 
	lowerthan(Len1,Len4).
\end{lstlisting}

\begin{figure}[h]
\centering
\includegraphics[width=9 cm]{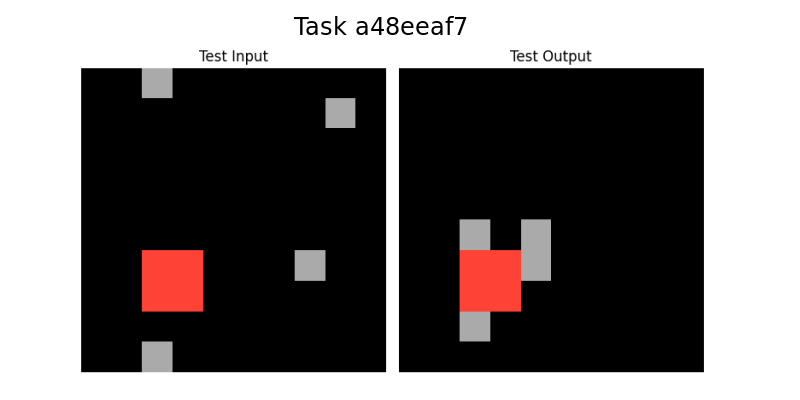}
\label{fig:a48eeaf7}
\end{figure}

\begin{lstlisting}[language=Prolog]
translate(Point1,Point2,X_dir,Y_dir):-
	member(Point1,Input_points),
	member(Rectangle1,Input_rectangles),
	point_straight_path_to(Point1,Rectangle1,X_dir,
		Y_dir,Orientation,Direction).

copy(Rectangle1,Rectangle2,Color_out,Clean_out):-
	member(Rectangle1,Input_rectangles),
	rectangle_attrs(Rectangle1,X1,Y1,X2,Y2,X3,Y3,X4,
		Y4,Color,Clean,Area),
	equal(Color_out,Color),
	equal(Clean_out,100).
\end{lstlisting}

\begin{figure}[h]
\centering
\includegraphics[width=9 cm]{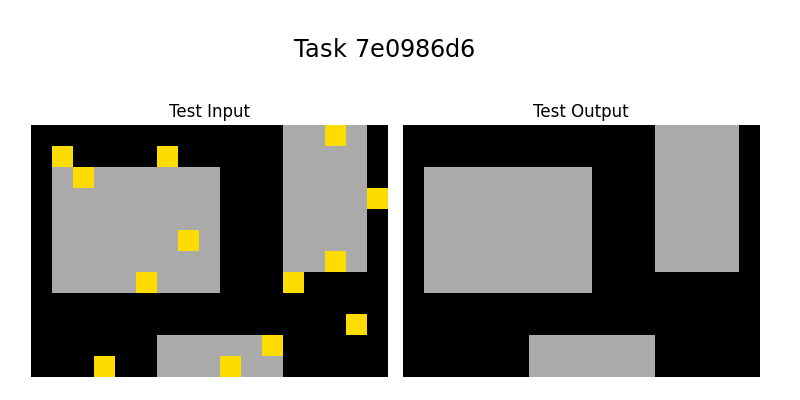}
\label{fig:7e0986d6}
\end{figure}

\begin{lstlisting}[language=Prolog]
copy(Rectangle1,Rectangle2,Color_out,Clean):-
	member(Rectangle1,Input_rectangles),
	rectangle_attrs(Rectangle1,X1,Y1,X2,Y2,X3,Y3,X4,
		Y4,Color,Clean,Area),
	equal(Color_out,Color),
	equal(Clean,100).
\end{lstlisting}

\begin{figure}[h]
\centering
\includegraphics[width=9.5 cm]{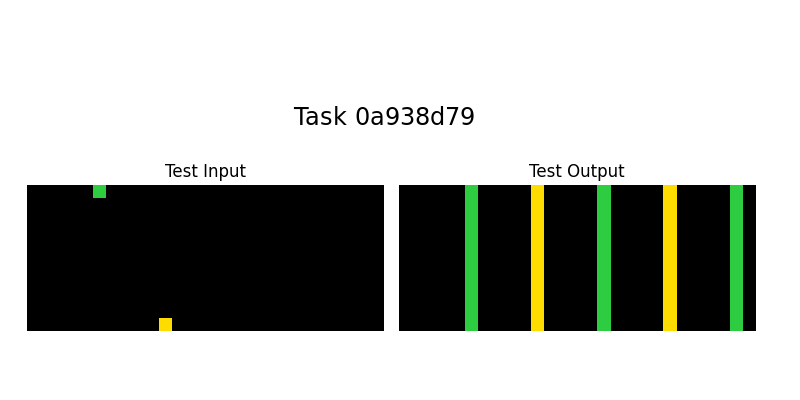}
\label{fig:0a938d79}
\end{figure}

\begin{lstlisting}[language=Prolog]
line_from_point(Point,Line,Len,Orientation,Direction):- 
	member(Point,Input_points), 
	equal(Len,X_dim), 
	equal(Orientation,'vertical').
	
translate(Line1,Line2,X_dir,Y_dir):-
	member(Line1,Input_lines),
	equal(X_dir,0),
	translate(Input_point1,Input_point2,X_dir,Y_dir),
	equal(Y_dir,2*Y_dir).
	
translate(Line1,Line2,X_dir,Y_dir):-
	member(Line1,Input_lines),
	equal(X_dir,0),
	translate(Input_point1,Input_point2,X_dir,Y_dir),
	equal(Y_dir,2*Y_dir).
	
translate(Line1,Line2,X_dir,Y_dir):-
	member(Line1,Input_lines),
	equal(X_dir,0),
	translate(Input_point1,Input_point2,X_dir,Y_dir),
	equal(Y_dir,2*Y_dir).
	
translate(Line1,Line2,X_dir,Y_dir):-
	member(Line1,Input_lines),
	equal(X_dir,0),
	translate(Input_point1,Input_point2,X_dir,Y_dir),
	equal(Y_dir,2*Y_dir).
\end{lstlisting}

\begin{figure}[h]
\centering
\includegraphics[width=9 cm]{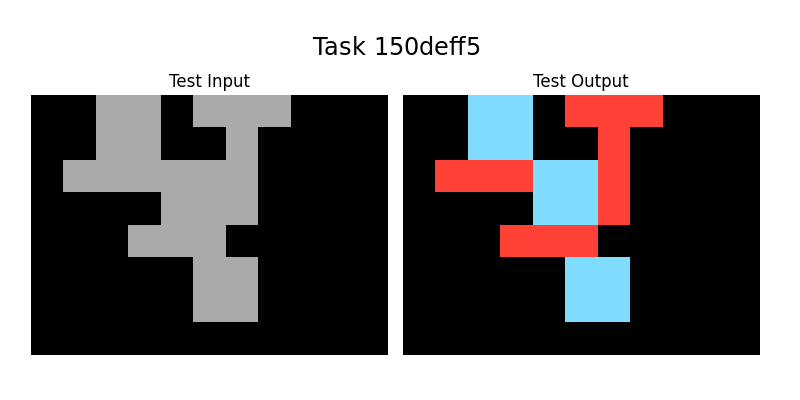}
\label{fig:150deff5}
\end{figure}

\begin{lstlisting}[language=Prolog]
copy(Rectangle1,Rectangle2,Color_out,Clear):-
	member(Rectangle1,Input_rectangles),
	rectangle_attrs(X1,Y1,X2,Y2,X3,Y3,X4,Y4,Color,
		Rectangle1,Clear,Area),
	equiv(Area,4),
	equal(Color_out,'blue'),
	equal(Clear,100).

copy(Line1,Line2,Color_out,Clear):-
	member(Line1,Input_lines),
	line_attrs(Line1,X1,Y1,X2,Y2,Color,Len,
		Orientation,Direction),
	equiv(Len,3),
	equal(Color_out,'red'),
	equal(Clear,100).
\end{lstlisting}


\end{document}